\documentclass[notitlepage,article,onecolumn,pre,aps,floatfix,nofootinbib]{revtex4-1}
\usepackage{amsmath}
\usepackage{amsthm}
\usepackage{amssymb}
\usepackage{mathrsfs}
\usepackage{makecell}
\usepackage{graphicx}
\usepackage{epstopdf}
\usepackage[usenames]{color}
\usepackage{xr}
\usepackage{placeins}
\usepackage[breaklinks=true,
            pdfborder={1 1 1},
            colorlinks=true,
            linkcolor=black,
            citecolor=blue,
            urlcolor=blue]{hyperref}
\usepackage[normalem]{ulem}
\usepackage{rotating}
\usepackage{verbatim}
\usepackage{bibentry}
\usepackage{xspace}
\usepackage[margin=1.0in]{geometry}
\usepackage{bm}
\usepackage{amssymb}
\usepackage{soul}
\usepackage{hyperref}
\usepackage{siunitx}
\usepackage{floatrow}
\usepackage[toc,page]{appendix}
\usepackage[table]{xcolor}
\usepackage{colortbl}
\usepackage{array}
\usepackage{xcolor}
\usepackage{fancyhdr}

\newcolumntype{C}[1]{>{\centering\arraybackslash}m{#1}}

\pdfoutput=1

\begin{document}

\title{Stacked tensorial neural networks for reduced-order modeling of a \\parametric partial differential equation}

\author{Caleb G. Wagner}
\email{caleb.wagner@parsons.us}
\email{c.g.wagner23@gmail.com}

\affiliation{Parsons Corporation, 14291 Park Meadow Drive, Suite 100
Chantilly, VA 20151}

\begin{abstract}
Tensorial neural networks (TNNs) combine the successes of multilinear algebra with those of deep learning to enable extremely efficient reduced-order models of high-dimensional problems. Here, I describe a deep neural network architecture that fuses multiple TNNs into a larger network, intended to solve a broader class of problems than a single TNN. I evaluate this architecture, referred to as a ``stacked tensorial neural network" (STNN), on a parametric PDE with three independent variables and three parameters. The three parameters correspond to one PDE coefficient and two quantities describing the domain geometry. The STNN provides an accurate reduced-order description of the solution manifold over a wide range of parameters. There is also evidence of meaningful generalization to parameter values outside its training data. Finally, while the STNN architecture is relatively simple and problem-agnostic, it can be regularized to incorporate problem-specific features like symmetries and physical modeling assumptions.
\end{abstract}

\maketitle
\thispagestyle{fancy}
\section{Introduction} \label{sec:introduction}
Partial differential equation (PDEs) lie at the core of many applications in scientific computing. Traditional numerical PDE algorithms, deriving from classical approximation theory, have proven highly effective and are widely used. Nevertheless, there are situations that pose challenges to traditional solvers. Three notable examples are \emph{multi-query}, \emph{real-time}, and \emph{slim} contexts. (This nomenclature is from Ref. \onlinecite{haasdonk2017reduced}.) \emph{Multi-query} problems require many repeated solves of a PDE at different parameter values or on different domains. Examples include inverse problems, uncertainty quantification, and design and optimization. \emph{Real-time} applications require that a PDE be solved in response to real-time changes in the environment. Finally, in \emph{slim} contexts, the available hardware is restricted to edge devices like phones or sensors. Traditional PDE algorithms -- though accurate and robust -- can be slow and resource intensive, which limits their applicability to these contexts.

One reason for this limitation is that classical methods do not scale well with problem size: computational cost grows rapidly with the number of independent variables and the grid resolution. This is called the \emph{curse of dimensionality}. On the other hand, it has been known for some time that \emph{physically relevant solutions} of PDEs, such as those generated by time evolution, tend to lie on or near a low-dimensional manifold that is roughly independent of the problem size \cite{Miranville2008,Foias1988,Robinson2001-lv}. In principle, approximating this manifold directly could bypass the curse of dimensionality. 

Recently, neural networks (NNs) have emerged as a promising method for constructing low-dimensional approximations to PDEs \cite{Han2018,Karniadakis2021,Faroughi2022,Goswami2023,Grohs_2023}. Here, the challenge is not so much finding a NN that solves a specific problem accurately, given initial and boundary conditions, but finding one that generalizes to contexts outside its training data -- for example, different domain geometries or PDE coefficients. This difficulty has hindered the use of NNs in the resource-constrained situations mentioned above, where it is impractical to retrain a NN once it has been deployed.

Toward this end, much recent work has focused on NN solvers for \emph{parametric PDEs} \cite{KHOO2020,Li2020GNO,li2020fourier,Geist2021,Lu2021,Wang2021,Cicci2022,FRESCA2022,Franco2023,Franco2023a,Romor2023}. Here, the goal is to find a single NN that solves a family of PDEs, whose members are generated by varying parameters such as the PDE coefficients. Empirically, deep neural networks have proven remarkably successful at solving parameterized PDE problems. Recent theoretical analysis correlates this success with the existence of a reduced-order basis defining the solution manifold, which allows the problem to be represented as a relatively small, parameterized linear system \cite{Kutyniok2021}. NNs, in turn, learn to invert this system as a function of its parameters. 

Here, I demonstrate a NN architecture, referred to as the stacked tensorial neural network (STNN), that performs well on a PDE with three independent variables and three varying parameters. The STNN is composed of a set of independent tensor networks, each of which takes the problem boundary conditions as input. The tensor networks may be thought of as local approximations of the aforementioned reduced-order linear system. To produce a continuous approximation over a suitable volume of parameter space, these networks are ``stacked" into a larger network, such that copies of the input are distributed across the tensor networks, processed, and combined at the end into a weighted sum (Fig. \ref{fig:STNN}).

The PDE was chosen as a prototypical example of a transport equation, which is a class of high-dimensional PDEs characterized by a coupling between spatial advection and internal degrees of freedom \cite{Duderstadt1979}. Transport equations are frequently treated using reduced-order models that seek to integrate over the internal degrees of freedom, eliminating them as independent variables \cite{Cercignani1990,Pomraning1992,McClarren2010,Kokhanovsky2019}. Developing such models often requires case-by-case analysis or additional modeling assumptions. By contrast, a well-designed NN could simplify this task by subsuming the problem-specific details into the training process.

Based on the results presented here, the STNN architecture is a plausible candidate for this type of reduced-order modeling as well as other parameterized PDE problems. Nevertheless, an issue is the need to generate labeled training data, which may be challenging for high-dimensional transport equations. I will revisit these broader aspects in section \ref{sec:discussion}, where I discuss avenues for future work.

 \section{Problem statement} \label{sec:problem-statement}
 We will solve the steady-state version of
 \begin{equation}
\frac{\partial f}{\partial t} + \ell \left( \cos w \cdot \frac{\partial f}{\partial x} +  \sin w \cdot \frac{\partial f}{\partial y} \right)  =  \frac{\partial^2 f}{\partial w^2},  \label{eq:pde-cartesian}
 \end{equation}
where $0 < w < 2\pi$, and $(x, y)$ are defined on a subset $D$ of $\mathbb{R}^2$. This equation has roots in Brownian motion and nonequilibrium statistical mechanics, where $f$ is the probability density of a set of underlying stochastic trajectories \cite{Marchetti2016,Wagner2017,Wagner2022}. It also has been applied in computer vision as a Brownian prior for identifying partially occluded edges \cite{Mumford1994}. The term on the right-hand-side describes diffusion in the angular coordinate $w$, and the left-hand-side describes advection along $\hat{\mathbf{u}} = (\cos w, \sin w)$. The parameter $\ell$ is a ``persistence length", which is the characteristic length governing spatial correlations in $f$.

Our focus is the steady-state density $\rho(x, y)$:
\begin{equation}
\rho(x, y) = \int f(x, y, w) \, dw.
\end{equation}
where $f(x, y, w)$ solves \eqref{eq:pde-cartesian} with $\partial_t f$ set to $0$. The domain $D$ is an annular region bounded by confocal ellipses. Let $a_1$ and $b_1$ be the minor and major axes of the inner ellipse, and $a_2$ and $b_2$ the same of the outer ellipse. A suitable rotation of the coordinate axes and rescaling of lengths will transform the inner ellipse so that $b_1 = 1$ and the long axis is vertically oriented. To simplify the network architecture, we require this step to be performed in preprocessing.
 
Let $\boldsymbol{\lambda}$ denote the problem parameters and $g$ the boundary data. Based on the above discussion, there are three parameters: $\boldsymbol{\lambda} = (\ell, a_1, a_2)$. Note that this setup includes circular boundaries as a special case ($a_1 = 1$). To formulate the boundary conditions, let $\hat{n}_1$ and $\hat{n}_2$ be the inward pointing normal vector along the inner and outer boundaries, respectively. For the problem to be well-posed, boundary conditions must be specified only where $\hat{n}_1 \cdot \hat{u} > 0$ or $\hat{n}_2 \cdot \hat{u} > 0$ \cite{Duderstadt1979,Wagner2019a}. Because $f(x, y, w)$ is a probability density, we also require that the boundary data is non-negative.

One of the main qualitative features of solutions of \eqref{eq:pde-cartesian} is a boundary layer that is non-analytic in $\ell$. There is also a long-range solution that can be expressed as an asymptotic series in $\ell$, and which appears when the net flux $\int \hat{u} f(r,\eta,w) dw$ is nonzero along the boundary \cite{Wagner2022}. We will see examples of this behavior in section \ref{sec:results}.

\subsection{Finite-difference representation}
The steady-state version of Eq. \eqref{eq:pde-cartesian} is solved by transforming from $(x,y,w)$ to polar coordinates $(r, \theta, w)$ or elliptical coordinates $(\mu, \eta, w)$, and applying a finite-difference discretization. The coordinate transforms are
\begin{align}
\begin{matrix}
x = r \cos \theta \\
y = r \sin \theta
\end{matrix} \, ; \qquad
\begin{matrix}
x = \sqrt{b_1^2 - a_1^2} \cosh \mu \cos\eta \\
y = \sqrt{b_1^2 - a_1^2} \sinh \mu \sin\eta
\end{matrix}
\end{align}
Denoting the inner and outer boundary conditions by $g_1$ and $g_2$, we have
\begin{equation}
\begin{matrix}
f(\mu = \mu_1, \eta, w) &= g_1(\eta, w) \quad \text{where  }\hat{n}_1 \cdot \hat{u} > 0 \\
f(\mu = \mu_2, \eta, w) &= g_2(\eta, w) \quad \text{where  }\hat{n}_2 \cdot \hat{u} > 0
\end{matrix} \label{eq:bc-general}
\end{equation}
and similarly for $(r,\theta)$. The finite-difference discretization is defined on a regular grid with dimensions $n_{r} \times n_{\theta} \times n_{w} = n_{\mu} \times n_{\eta} \times n_{w} = 256 \times 64 \times 32$. In the $\eta$ ($\theta$) and $w$ directions, the grid spacing is uniform, while in the $\mu$ ($r$) direction, the grid points are concentrated near the inner and outer boundaries, which is required to resolve the boundary layer. If $\nu$ denotes $n_{r}+2$ points spaced uniformly between $-\pi/2$ and $\pi/2$, then the $r$-grid ($n_{r}$ interior nodes and 2 boundary nodes) is
\begin{equation}
r = (b_2 - b_1) (\sin \nu + 1) / 2 + b_1.
\end{equation}
To ensure that the $\mu$-grid transforms continuously into the $r$ grid as the minor axis approaches $1$, the $\mu$ grid is scaled logarithmically:
\begin{equation}
\log \mu = (e^{\mu_2} - e^{\mu_1}) (\sin \nu + 1) / 2 + e^{\mu_1}.
\end{equation}
Eq. \eqref{eq:pde-cartesian} is discretized using fourth-order central differences for $\partial^2 / \partial w^2$ and first-order upwind differences for $\mu$ and $\eta$ \cite{Zwillinger2021}. To simplify notation, the remainder of this paper uses $\mu$ and $\eta$, with the analogous relations implied for $r$ and $\theta$. 

The discretized linear system was solved iteratively using the Generalized Minimum RESidual (GMRES) algorithm, as implemented in the CUDA-accelerated Python library CuPy \cite{cupy_learningsys2017,cupy_github}. No preconditioner was used. Instead, the solution operator was constructed explicitly at selected values of $\boldsymbol{\lambda}$ (see Eq. \eqref{eq:AB-inv}) and used to generate initial guesses for GMRES, which resulted in reliable convergence for the problem inputs considered here.

\subsection{Statement of the discretized problem} \label{subsec:discretized-problem}
Because the STNN operates on the discretized version of \eqref{eq:pde-cartesian}, we use a separate notation for the gridded representation of the problem variables:
\begin{align}
f(x, y, w) &\rightarrow \mathbf{f} \quad \text{($n_{\mu} n_{\eta} n_w$-dimensional vector)} \\
\rho(x, y) &\rightarrow \boldsymbol{\rho} \quad \text{($n_{\mu} n_{\eta}$-dimensional vector)} \\
g(\eta, w) &\rightarrow \mathbf{g} \quad \text{($n_{\eta} n_w$-dimensional vector)}
\end{align}
Here, $\mathbf{g}$ combines $g_1$ and $g_2$ (Eq. \eqref{eq:bc-general}) into a single vector. For fixed $\boldsymbol{\lambda}$, Eq. \eqref{eq:pde-cartesian} is linear, so the solution can be written as
\begin{equation}
\mathbf{f} = A_{\boldsymbol{\lambda}} \cdot \mathbf{g};\qquad
\boldsymbol{\rho} = B_{\boldsymbol{\lambda}} \cdot \mathbf{g}, \label{eq:AB-inv}
\end{equation}
where $A_{\boldsymbol{\lambda}}$ and $B_{\boldsymbol{\lambda}}$ are dense, $\boldsymbol{\lambda}$-dependent matrices with shapes $(n_{\mu} n_{\eta} n_{w}) \times (n_{\eta} n_{w})$ and $(n_{\mu} n_{\eta}) \times (n_{\eta} n_{w})$.

The objective of the STNN architecture is to approximate $A_{\boldsymbol{\lambda}}$. That is, the STNN seeks a map $\mathscr{N}$ from $(\boldsymbol{\lambda}, \mathbf{g})$  to the steady-state density $\boldsymbol{\rho}$:
 \begin{equation}
\mathscr{N} : \left(\boldsymbol{\lambda}, \mathbf{g} \right) \rightarrow \boldsymbol{\rho}. \label{eq:stnn-objective}
 \end{equation}

\section{Network architecture} \label{sec:network-architecture}
\begin{figure}
    \centering
    \includegraphics[width=.99\textwidth]{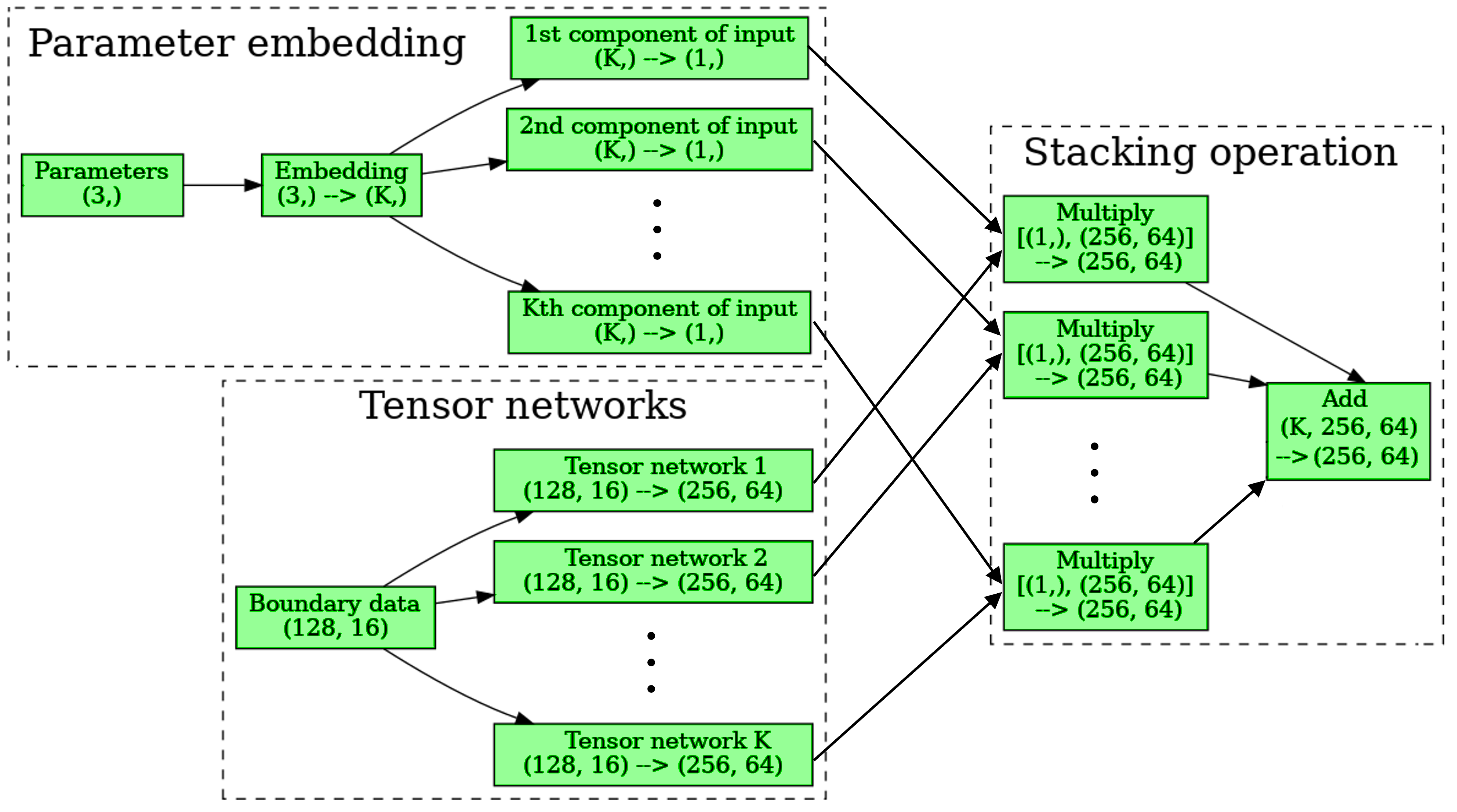}
\caption{The stacked tensorial neural network (STNN) architecture. The array shapes correspond to the problem from section \ref{sec:problem-statement}. In particular, the parameter embedding takes $\boldsymbol{\lambda} = (\ell, a_1, a_2)$ as input, while each tensor network takes the boundary data $\mathbf{g}$ as input. The output of the stacking operation, and the network as a whole, is the finite-difference representation of the density $\boldsymbol{\rho}$.}
    \label{fig:STNN}
\end{figure}
Here, I describe the STNN architecture as applied to the problem from the previous section. First, note that for fixed $\boldsymbol{\lambda}$, $\boldsymbol{\rho}$ can be exactly predicted from the boundary data $\mathbf{g}$ using a single-layer dense network with linear activation: the bias is 0, and the weights are the matrix elements of $B_{\boldsymbol{\lambda}}$, Eq. \eqref{eq:AB-inv}. However, there is no easy way to express the elements of $B_{\boldsymbol{\lambda}}$ as a function of $\boldsymbol{\lambda}$. Even for fixed $\boldsymbol{\lambda}$, the network has $\sim\!30$ million parameters, making it difficult to efficiently incorporate into a higher-level architecture. We address these difficulties in two steps.

\vspace{3mm}

\textbf{Step 1: tensor networks}

Tensor networks are feed-forward networks expressing a sequence of tensor contraction operations \cite{Reyes2021,Wang2021,Wang2023}. They are inspired by tensor decompositions in multilinear algebra \cite{Kolda2009,Oseledets2011,Cichocki2016,Brunton2019}. In our case, the idea is to replace the single layer dense network
\begin{equation}
\boldsymbol{\rho} = B_{\boldsymbol{\lambda}} \cdot \mathbf{g} 
\end{equation}
with the $M$-fold tensor network
\begin{equation}
\boldsymbol{\Pi} = \mathbf{Q}_1 \cdot \mathbf{Q}_2 \cdots \mathbf{Q}_M \cdot \mathbf{G}, \label{eq:general-tensor-decomp}
\end{equation}
where $\cdot$ denotes an appropriate contraction, and $\boldsymbol{\Pi}$ and $\mathbf{G}$ are reshaped (``tensorized") versions of $\boldsymbol{\rho}$ and $\mathbf{g}$. 

There are many tensor decompositions defined in the literature (see Ref. \onlinecite{Wang2023} for a review); here, we use the \emph{tensor train decomposition}, which has been used to construct extremely efficient tensorized solution algorithms for high-dimensional PDEs \cite{Kormann2015,Matveev2016,Dolgov2012,Manzini2023,truong2023tensor,kiffner2023tensor,ye2023quantized}. Under constraints on the ranks of the constituent tensors (called \emph{cores}), the tensor train decomposition can be computed as a nested sequence of truncated singular value decompositions. However, keeping in mind our goal of \eqref{eq:stnn-objective}, we broaden the interpretation by treating the tensor cores as trainable parameters. Then, we arrive at the following \emph{tensor train layer} (TTL) \cite{novikov2015tensorizing}:
\begin{equation}
\boldsymbol{Y}\left(i_1, \ldots, i_d\right)=\sum_{j_1, \ldots, j_d} \boldsymbol{Q}_1\left[i_1, j_1\right] \ldots \boldsymbol{Q}_d\left[i_d, j_d\right] \mathbf{X}\left(j_1, \ldots, j_d\right) \label{eq:TTL-general}
\end{equation}
Here, $\boldsymbol{Q}_k[i_1,j_1]$ is a matrix with dimensions $r_k \times r_{k-1}$. The $r_k$ are called the ``ranks" and must satisfy $r_0 = 1$ and $r_d = 1$, while indices $i_k, j_k$ have range $i_k \in [1, L_k]$ and $j_k \in [1, U_k]$. Both $L_k$ and $U_k$, as well as $d$, are constrained by the input and output dimensions of the layer, while the ranks can be chosen more freely. In our case, we use $d = 7$ and $L_{1 \ldots d} =  (4, 4, 4, 4, 4, 4, 4)$, 
while $U_{1 \ldots d}$ and $r_{0 \ldots d}$ are treated as tunable hyperparameters (see Table \ref{tab:hyperparameter_trials} for examples).

Comparing Eqs. \eqref{eq:general-tensor-decomp} and \eqref{eq:TTL-general}, we see that the ``default" choice for the layer input and output is $\mathbf{g}$ and $\boldsymbol{\rho}$, respectively (technically, their tensorized versions $\mathbf{G}$ and $\boldsymbol{\Pi}$). While this configuration is included in our model evaluation (section \ref{sec:results}), it does not easily incorporate symmetries or physical constraints. To address this limitation, we focus on models with an additional preprocessing layer, such that the input to the TTL is not $\mathbf{g}$ but $\mathbf{\bar{g}}$, defined as follows:
\begin{align}
[\mathbf{\bar{g}}]_i &\equiv \sum_{j=1}^{n_{\eta}}[\mathbf{h}^+]_j [\mathcal{P} \mathbf{g}]_{ij} + \sum_{j=n_{\eta}+1}^{2n_{\eta}}[ \mathbf{h}^-]_{j-n_{\eta}} [\mathcal{P} \mathbf{g}]_{ij} \label{eq:einsum-layer} \\
&\simeq \int_{\hat{n}_1\cdot\hat{u}>0} h^+(\phi) g_1(\eta, \eta - \phi) d\phi + \int_{\hat{n}_2\cdot\hat{u}>0} h^-(\phi) g_2(\eta, \eta - \phi) d\phi \\
&\quad\text{where } \phi \equiv \eta - w. \nonumber
\end{align}
Here, $\mathcal{P}$ is a permutation operator that transforms from $(\eta, w)$ to $(\eta, \phi)$ coordinates in finite-difference space, while $\mathbf{h}^+$ and $\mathbf{h}^-$ are vectors containing $n_{\eta}$ trainable weights each. The layer defined by \eqref{eq:einsum-layer} regularizes the network by eliminating direct connections across the $w$-axis. Contracting over $\phi$ instead of $w$ also enforces a certain rotational symmetry by requiring that the weights depend only on the angle with respect to the local unit normal vector ($\phi$). This constraint is not strictly satisfied when the domain boundaries have non-constant curvature, but the loss of accuracy appears to be outweighed by the benefits of regularization.

A preliminary hyperparameter search indicated that the above layer was a bottleneck on the network's representational capacity. As a result, a modified version was also used, containing additional, independent weight vectors $\mathbf{h}^{\pm}$. This modification is in the spirit of moment approximations in physics, in which a distribution function $f(\mathbf{r}, \mathbf{p})$ is projected onto a finite set of moments, each of which is an integral of $h(\mathbf{p}) f(\mathbf{r}, \mathbf{p})$ over $\mathbf{p}$, for some weight function $h(\mathbf{p})$ (e.g.,  \cite{Widder1989,Fan2019,Viehland2018}).

Eqs. \eqref{eq:TTL-general} and \eqref{eq:einsum-layer} define the tensor networks comprising the STNN. For fixed $\boldsymbol{\lambda}$, replacing $B_{\boldsymbol{\lambda}}$ with a tensor network reduces the number of trainable parameters from $\sim \!3 \times 10^{6}$ to $\sim \!10^4$ while maintaining good accuracy.

\vspace{3mm}

\textbf{Step 2: stacking the tensor networks}

\vspace{1mm}

Second, we stack $K$ individual tensor networks into a larger network $\mathscr{N}(\boldsymbol{\lambda}, \mathbf{g})$ designed to model the $\boldsymbol{\lambda}$-dependence of the problem (Fig. \ref{fig:STNN}). The key feature is a sequence of layers that takes $\boldsymbol{\lambda}$ as input and outputs the coefficients of a weighted average over the tensor networks. To achieve the representational capacity necessary for modeling the $\boldsymbol{\lambda}$-dependence of the problem, $\boldsymbol{\lambda}$ is first embedded in a space with dimension $N_E = 30$, then passed through multiple dense layers with ReLu activation. In the last step, the $N_E$-dimensional vector is reduced to $K$ dimensions and fed into a softmax layer. The output is a set of $K$ numbers that sum to $1$, which are used to form a linear combination of the tensor networks.

\vspace{3mm}

\textbf{Normalization}

\vspace{1mm}

Like the solution operator of the PDE, the tensor networks are linear in their input $\mathbf{g}$ and commute with scalar multiplication, so their inputs and outputs do not require normalization. For training, however, applying normalization helps balance the loss equally among the simulation instances and may improve numerically stability, so the training reported here normalized $\mathbf{g}$ with respect to the average absolute value of $\boldsymbol{\rho}$. For very small or large $\mathbf{g}$, normalization may also be required to avoid overflow or underflow during inference; however, this did not appear to be necessary for the simulations considered here. For the embedding layer, the inputs $\boldsymbol{\lambda}$ are normalized with respect to the upper and lower limits of the training data; see Eq. \eqref{eq:training-data-volume} in section \ref{sec:implementation-training} below.

\subsection{Implementation and training} \label{sec:implementation-training}
The STNN was implemented and trained in TensorFlow v2.6.2 \cite{tensorflow2015-whitepaper}, and the tensor network \eqref{eq:TTL-general} was represented using the TensorFlow-compatible library t3f \cite{t3f_github,Novikov2020}.

 Before training, we need to define the space of inputs we would like the network to learn. For the boundary conditions $g(\eta,w)$, corresponding to the $n_{\eta} n_w$-dimensional vector $\mathbf{g}$, the most direct approach would be to choose a set of values for $\boldsymbol{\lambda}$, and for each of these, minimize an appropriate loss function over a $n_{\eta} n_w$-dimensional basis for $\mathbf{g}$. However, this objective  encompasses inputs that essentially amount to noise, and there is evidence that the STNN lacks the representational capacity for arbitrary inputs. This issue is revisited in sections \ref{sec:results-2} and section \ref{sec:discussion-1}. For now, we restrict attention to inputs whose spatial frequency is not too large, so they are adequately covered by the finite-difference grid. A straightforward way to incorporate this constraint is to build $\mathbf{g}$ from a Fourier series with predefined decay rates for the coefficients. Toward this end, we define the following ``function generators":
 \begin{align}
\text{fgen}_{i}(\eta, \phi) &= \frac{a_0}{2} +  \sum_{n,m} \left( a_{nm} \cos n\eta + b_{nm} \sin n\eta \right) \left( c_{nm} \cos m \phi + d_{nm} \sin m \phi \right) \label{eq:fgen-defs}, \\
&\quad \text{where } \phi = \eta - w, \,\,a_{nm} = q_i(n, m) \cdot \mathrm{U}[-1, 1] \text{  and  same for } b_{nm}, c_{nm}, d_{nm}. \nonumber
 \end{align}
 Here, $a_0$ is chosen to be the smallest value that makes the given function nonnegative, $\mathrm{U}[-1, 1]$ denotes the uniform distribution between $-1$ and $1$, and $q_i(n, m)$ is the coefficient decay function. We consider two types of decay functions:
 \begin{align}
&q_i(n, m) = (n m)^{-i/2}, \quad i \geq 0 \\
&q_{\text{exp}}(n, m) = \exp(-s_{11} n^2 - s_{22} m^2 - 2 s_{12} n m), \quad s_{11}, s_{22}, s_{12} > 0.
 \end{align}
Training and validation data were generated using $ \text{fgen}_{\text{exp}}$ and $\text{fgen}_i$ for $i = 0, 1, 2, 4$. For each sample, 12 terms were included in the sum \eqref{eq:fgen-defs}, with $n$ and $m$ drawn from $(2, n_w)$. To increase the diversity of the dataset, a fraction of samples were also multiplied by a damping function designed to set the value to $0$ on parts of the domain. As for the training data, $\boldsymbol{\lambda}$ was sampled from the following volume:
\begin{equation}
0.01 < \ell < 100; \qquad 2 < a_2 < 20; \qquad 0.2 < a_1 \leq 1 \,. \label{eq:training-data-volume}
\end{equation}
In the end, the training and validation consisted of 27108 and 3013 samples.

Note that the validation data is used mainly to estimate model performance during training and make adjustments as needed. For the \emph{test data}, it is essential to use function generators that are different from the ones used for training and validation; doing so reduces the risk that the network merely learns spurious patterns of the generators in its training data. Here, we use three distributions for our test data. \textbf{Test group 1} ($N = 2647$) is generated from random piecewise linear functions, where both the number of pieces and the slopes are randomly chosen. \textbf{Test group 2} ($N = 6$) contains ``zero-flux" solutions, in which the net flux across the inner boundary is zero. Because density is conserved, the net flux at the outer boundary must also be zero, which implies an extra constraint in the problem. This constraint can be satisfied by setting the boundary condition at the outer boundary to an appropriate constant. Such constrained solutions are unlikely to appear consistently in the training data, so they are a good way to check for overfitting. Finally, \textbf{test group 3} ($N = 2680$) consists of samples where $\boldsymbol{\lambda}$ falls outside the range covered by training data. Specifically, the boundary conditions are generated using $\text{fgen}_{11}$, while $\boldsymbol{\lambda}$ is sampled from $100 < \ell < 200$, $1.5 < a_2 < 50$, and $0.2 < a_1 \leq 1$.
 
Training used the Adam optimizer with a batch size of 128, an adaptive learning rate, and mean-squared error as the loss. The learning rate was initialized at 0.005 and updated using the TensorFlow callback ReduceLROnPlateau, which halved the learning rate whenever the validation loss plateaued for more than 20 epochs. Overall, the training was performed in two passes. First, the layer \eqref{eq:einsum-layer} was supplemented with a regularization penalizing high-frequency components of the weight vectors $\mathbf{h}^{\pm}$, and 200 epochs were performed. Second, the regularization was removed, the learning rate reset to 0.002, and training continued for at least 200 epochs. The trainable parameters were initialized using the TensorFlow and t3f defaults: namely, glorot uniform and zero bias for the dense layers, and glorot normal for the tensor train layers \cite{pmlr-v9-glorot10a,t3f_github}.

\begin{table}
\centering
\begin{tabular}
{cC{0.6cm}C{0.6cm}C{0.8cm}C{3.0cm}C{1.55cm}C{1.6cm}C{1.58cm}C{1.58cm}C{1.58cm}}
\toprule
\noalign{\vskip-16pt}
\thead{\\\\\textbf{Trial}} & \thead{\\\\$\boldsymbol{K}$} & \thead{\\\\$\boldsymbol{d}$}  & \thead{\\\\$\boldsymbol{W}$} & \thead{\\\\\\\textbf{Tensor network}\\ \textbf{ranks}*} & \thead{\\\\$\boldsymbol{N_{\mathrm{params}}}$} & \multicolumn{4}{c}{\thead{\textbf{Relative $L^2$ norm difference (sample average)}}}  \\
\noalign{\vskip-16pt}
\cline{7-10}
& & & & & & \thead{\textbf{Train} \\ ($N\!=\!27108$)} & \thead{\textbf{Test 1} \\ ($N \!=\!2647$)} & \thead{\textbf{Test 2} \\ ($N \!=\!6$)} & \thead{\textbf{Test 3} \\ ($N \!=\!2860$)}  \\  
\hline
\noalign{\smallskip}

1 & 10 & 10 & 1 & \makecell{$(1, 7) + (16)_4 + (7, 1)$} & 87580 & 0.018 & 0.014 & 0.013 & 0.028\\

2 & 20 & 10 & 1 & \makecell{$(1, 7) + (16)_4 + (7, 1)$} & 168530 & 0.015 & 0.012 & 0.012 & 0.026\\

3 & 30 & 10 & 1 & \makecell{$(1, 7) + (16)_4 + (7, 1)$} & 249480 & 0.014 & 0.011 & 0.013 & 0.031\\

\noalign{\vskip5pt}

4 & 10 & 10 & 2 & \makecell{$(1) + (16)_5 + (7, 1)$} & 101260 & 0.020 & 0.013 & 0.012 & 0.028\\

5 & 20 & 10 & 2 & \makecell{$(1) + (16)_5 + (7, 1)$} & 195890 & 0.014 & 0.011 & 0.017 & 0.027\\

6 & 30 & 10 & 2 & \makecell{$(1) + (16)_5 + (7, 1)$} & 290520 & 0.013 & 0.010 & 0.010 & 0.022\\

\noalign{\vskip5pt}

7 & 10 & 10 & $\text{N/A}^\dagger$ & \makecell{$(1, 7) \mathop{+} (16)_4 \mathop{+} (7, 1)$} & 137900 & 0.017 & 0.015 & 0.015 & 0.028\\

\noalign{\vskip4pt}
\botrule
\end{tabular}
\flushleft
\vspace{-5pt}
$\qquad^* (c)_k$ is shorthand for $c$ repeated $k$ times. For example, $(a, b) + (c)_3$ represents $(a,b,c,c,c)$.\\ 
$\qquad^\dagger$The preprocessing layer \eqref{eq:einsum-layer} was not used here.                               
\caption{Trials of different hyperparameter combinations for the stacked tensorial neural network (STNN). Here, $K$ is the number of tensor networks, $d$ is the depth of the embedding layer, and $W$ is the number of weight vectors in the preprocessing layer, Eq. \eqref{eq:einsum-layer}. The accuracy of the STNN is measured on three test datasets, which are described in section \ref{sec:results}.}
\label{tab:hyperparameter_trials}
\end{table}

\newpage
\section{Results} \label{sec:results}

 \begin{figure}
    \centering
    \includegraphics[width=.8\textwidth]{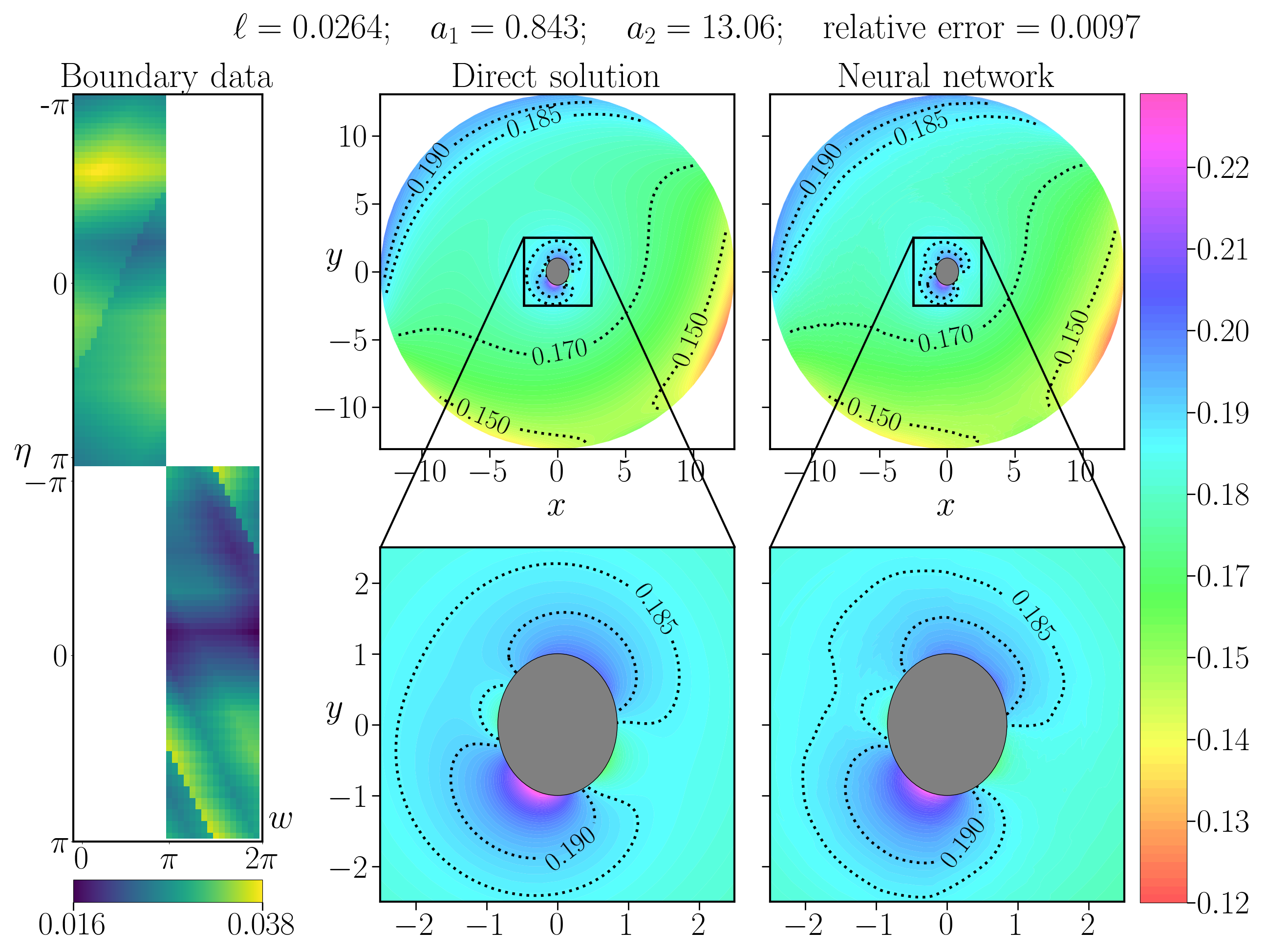}
    
    \includegraphics[width=.8\textwidth]{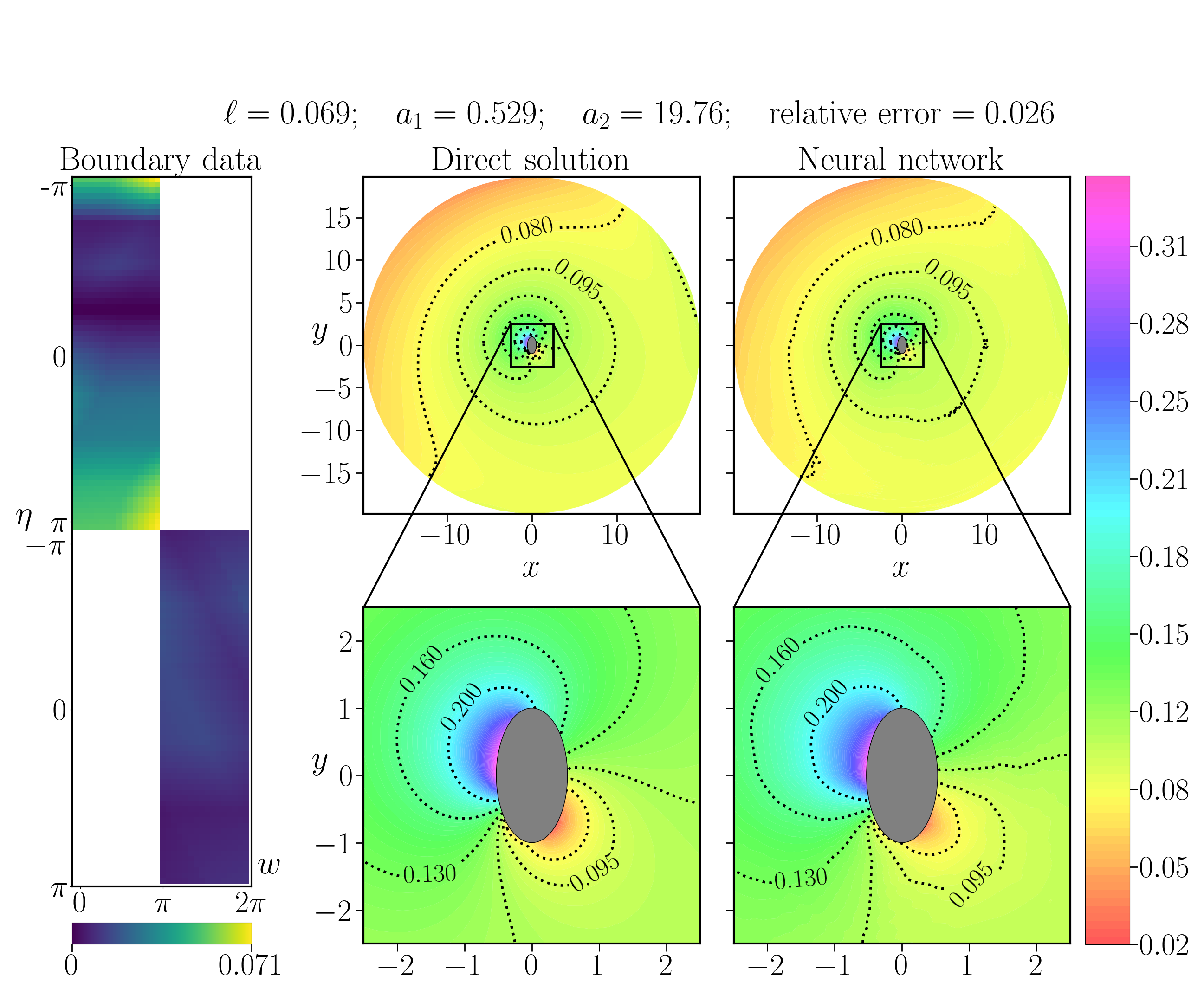}
\caption{\footnotesize{Comparisons between the direct solution and STNN prediction for two examples from test group 1. The STNN corresponds to trial 5 in Table \ref{tab:hyperparameter_trials}. For each example, the leftmost plot shows the boundary data on the inner and outer boundaries (Eq. \eqref{eq:bc-general}), while the remaining plots show the density $\rho(x,y)$. The top example is near the median with respect to the relative $L^2$ norm difference, while the bottom is near the 94th percentile, i.e., 94\% of the simulations in test group 1 (2647 total) have smaller relative error.}}
    \label{fig:validation-tests}
\end{figure}

Table \ref{tab:hyperparameter_trials} summarizes the results of training for different hyperparameter combinations. The examples in this section use the network from trial 5.

\subsection{Accuracy on test data} \label{sec:results-1}

The primary measure of accuracy is the relative $L^2$ error, $\epsilon = ||\boldsymbol{\rho}_{\text{STNN}} - \boldsymbol{\rho}|| / ||\boldsymbol{\rho}||$. Across \textbf{test group 1} (2647 simulations), the average, median, and maximum of $\epsilon$ were 0.011, 0.0091, and 0.121, respectively. Fig. \ref{fig:validation-tests} gives a visual comparison of the prediction and ground truth for two examples from the test dataset, corresponding to the median and 94th percentile with respect to $\epsilon$.

\begin{figure}
    \centering
    \includegraphics[width=.9\textwidth]{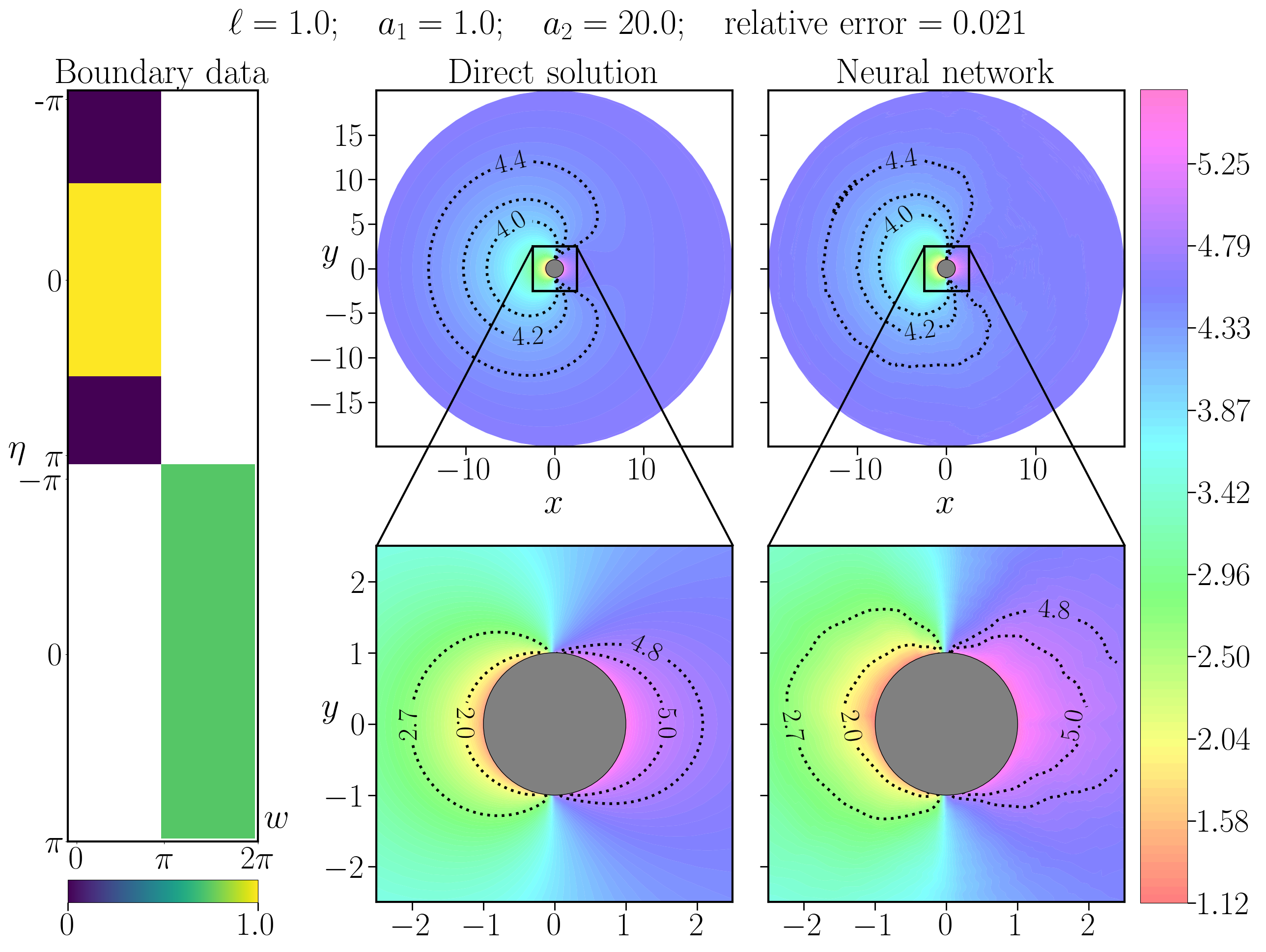}
\caption{Test of the stacked tensorial neural network (STNN) on a ``zero-flux" solution. Such solutions satisfy an extra constraint on the boundary conditions that is not specifically built into the training data. The leftmost plot shows the boundary data on the inner and outer boundaries (Eq. \eqref{eq:bc-general}), while the remaining plots show the density $\rho(x,y)$. The relative $L^2$ error is about 2.1\%.}
    \label{fig:zero_flux_example}
\end{figure}

Fig. \ref{fig:zero_flux_example} shows the results for one of the zero-flux problems from \textbf{test group 2}, corresponding to a step function boundary condition in the circular geometry. Here, the boundary data equals 1 on the right half of the inner boundary and 0 on the left half. The STNN prediction has a relative error of $2.1 \%$ and seems to correctly handle the zero-flux constraint.

\begin{figure}
    \centering
    \includegraphics[width=.9\textwidth]{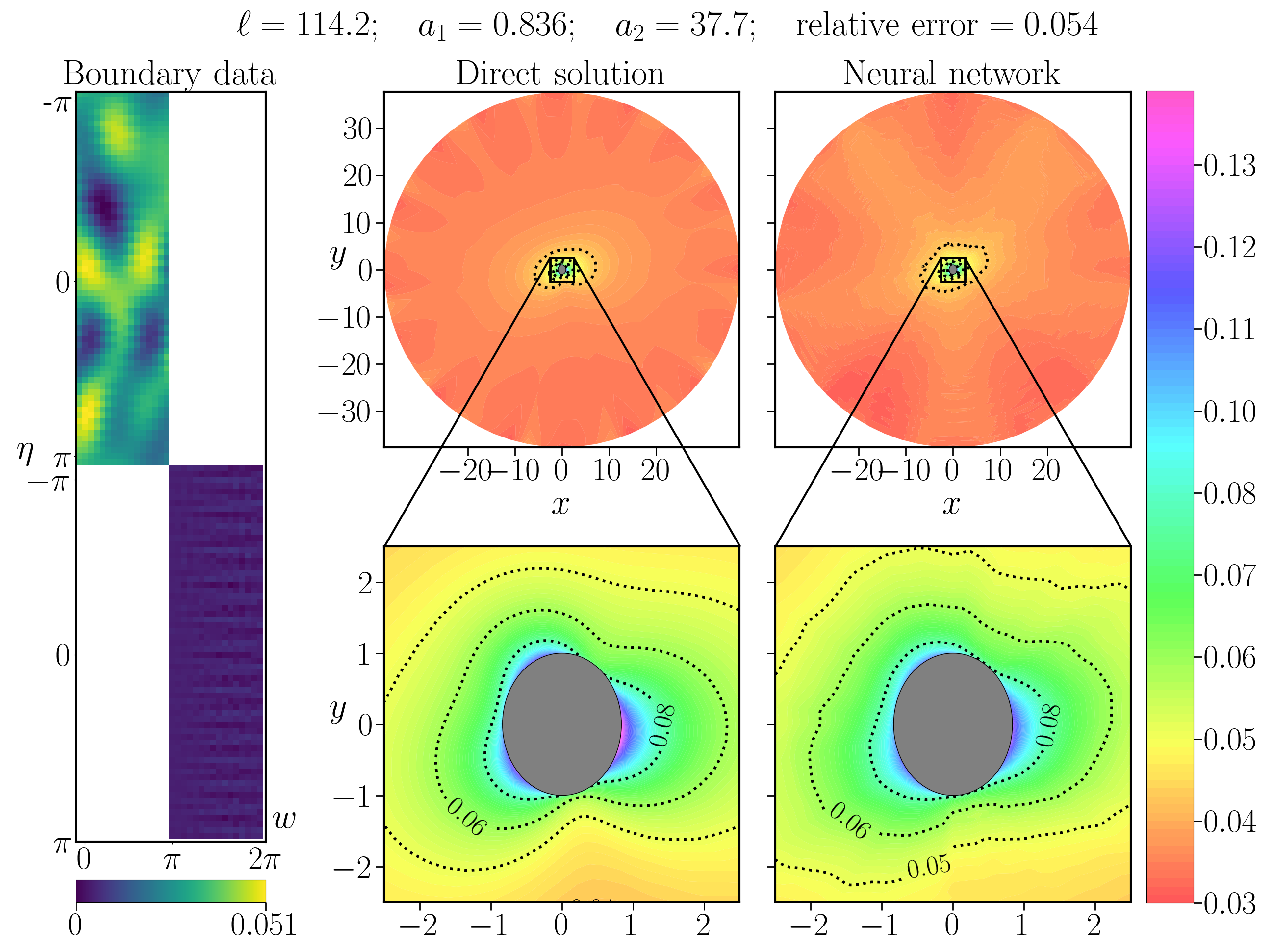}
\caption{Test of the STNN for parameter values $(\ell, a_1, a_2)$ outside the range covered by the training data. This example is taken from a set of 2860 randomly generated problem inputs, as described in section \ref{sec:results}. The leftmost plot shows the boundary data on the inner and outer boundaries (Eq. \eqref{eq:bc-general}), while the remaining plots show the density $\rho(x,y)$. This example is near the 92nd percentile with respect to the relative error, i.e. 92\% of the 2860 simulations have smaller relative error.}
    \label{fig:outofrange_params}
\end{figure}

Finally, we consider \textbf{test group 3}, in which $\boldsymbol{\lambda}$ is sampled from outside the range used for training (section \ref{sec:implementation-training}). Here, the median and 95th percentile relative errors were 0.020 and 0.054, while the maximum was 0.27 (see Fig. \ref{fig:outofrange_params}). The median error is modest, only twice the median error when $\boldsymbol{\lambda}$ is drawn from the range covered by the training data. This suggests that the STNN is learning the actual $\boldsymbol{\lambda}$-dependence of the solution manifold.

\subsection{Limits of applicability} \label{sec:results-2}

Understanding the situations that ``break" the STNN is important from both a practical and fundamental point of view. Here, I show two categories of inputs that seem to fall outside the representational capacity of the STNN; in other words, adjusting the model hyperparameters or increasing the training data is unlikely to significantly improve performance on these cases.

\begin{figure}
    \centering
    \includegraphics[width=.99\textwidth]{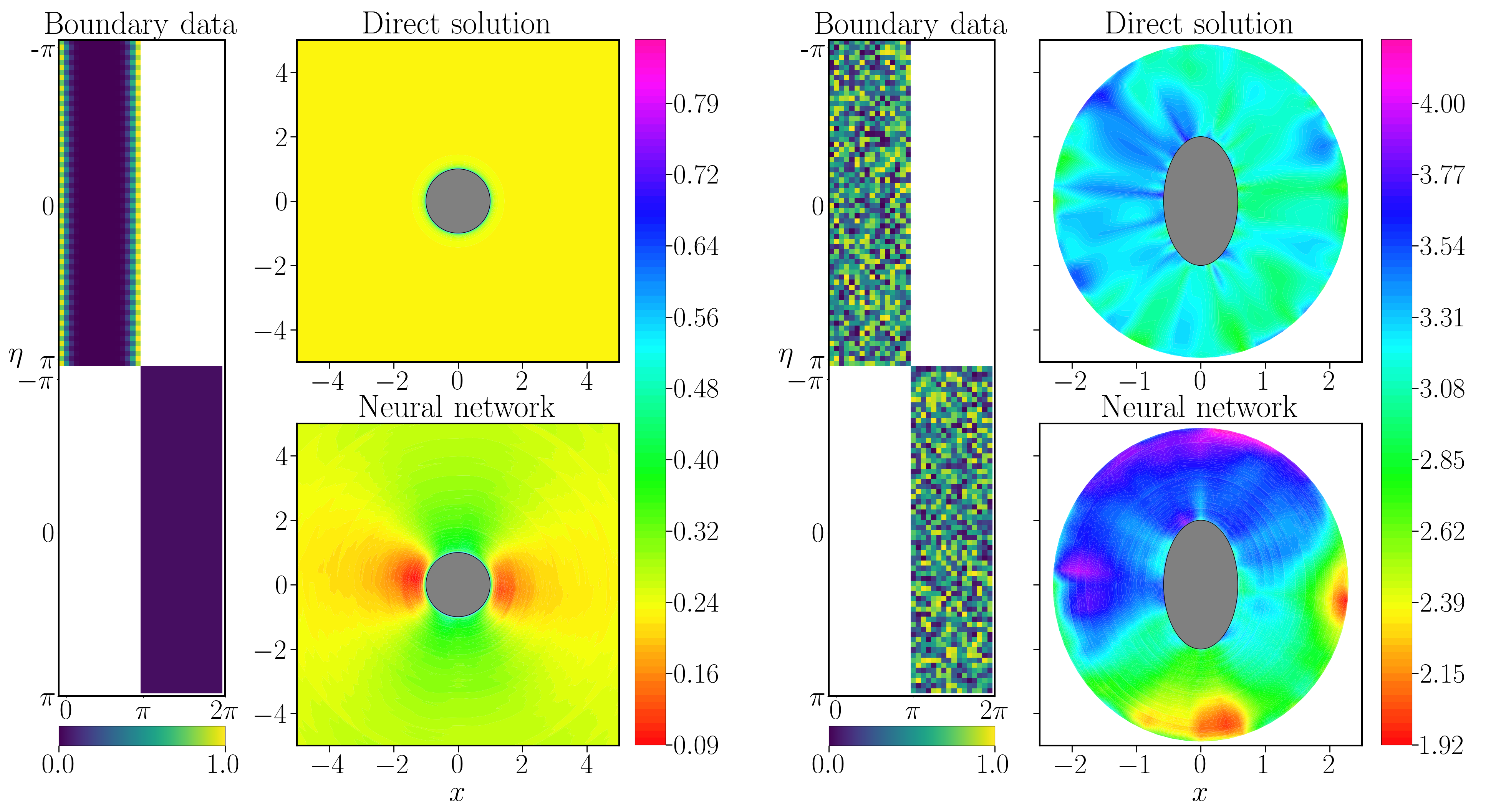}
\caption{Illustration of two types of boundary conditions that appear to fall outside the representational capacity of the STNN. On the left is a spherically symmetric boundary layer solution, which is nonzero only very close to the boundary. The example on the right corresponds to spatially uncorrelated boundary data, where the value of each boundary node is sampled uniformly from $[0,1]$. Sections \ref{sec:results-2} and \ref{sec:discussion-1} examine the implications of these examples for understanding the representational capacity of the STNN.}
    \label{fig:outside_rep_cap}
\end{figure}

The first category is spatially uncorrelated noise. Here, each element in the boundary data $\mathbf{g}$ is sampled uniformly from the range $[0, 1]$. Figure \ref{fig:outside_rep_cap} shows an example. Due to the linearity of the tensor networks, the STNN prediction has the right order of magnitude, but fails to capture any other features of the direct solution. On the other hand, if the correlation length of $\mathbf{g}$ is increased to about twice the grid resolution, the STNN predictions improve significantly. Whether this apparent length scale limitation is tied to an intrinsic length or merely the finite-difference grid resolution is not clear and would require further tests at different grid resolutions.

The second category of inputs corresponds to certain \emph{boundary layer solutions}, which are characterized in detail in Ref. \cite{Wagner2022}. These solutions are nonzero only within a very small distance $\epsilon \sim 0.1 \ell$ from the boundary. They are present in the general solution of the problem but are not important when large-scale fluxes are present. However, if the boundary data is a function of $\phi = \eta - w$ only, $g(\theta, w) = \bar{g}(\phi)$, these are the only solutions present, aside from an additive constant. The width of the boundary layer is especially small when $\bar{g}(\phi)$ is symmetric about $\phi = 0$ and concentrated near $\phi = \pm \pi/2$ \cite{Wagner2022,WagnerDissertation}. Figure \ref{fig:outside_rep_cap} shows an example of this type, where $\bar{g}(\phi) = \exp(-10 \sin^2 \phi)$. The STNN manages to predict some of the boundary layer, but the width and magnitude are not quite right, and spurious spatial variations are visible elsewhere in the domain. The relative error is $0.147$. We will revisit these examples in section \ref{sec:discussion-1}, where the representational capacity of the STNN is discussed in more detail.

\subsection{Performance benchmarks} \label{sec:results-3}
To assess the potential performance benefits of the STNN, we should compare it with the ``best performing" classical method. Unfortunately, the latter depends on the problem at hand and the available computational hardware. Even if these factors are controlled, there can still be significant variations in performance. For example, the GMRES algorithm used here is an iterative method whose performance strongly depends on the preconditioning (if any) and the initial guess for the solution. Finally, performance can be measured in more than one way: for example, computational time, memory overhead, or energy usage.

Keeping in mind these caveats, we will compare the \emph{computational time} of the STNN with GMRES, using either a GPU (NVIDIA RTX 3090 Ti) or CPU (Intel i9-10900KF) with 8 openMP threads. We will assume the GMRES algorithm has been supplied with a good initial guess, obtained from the solution at a nearby set of parameters. By averaging over trials with different problem inputs, we find that the average execution time for GMRES is $4.66$ seconds on the GPU and $51.9$ seconds on the CPU.\footnote{These values were based on 360 simulations for the GPU and 29 for the CPU. In the latter case, the GMRES iterations stalled at the 30th simulation, so only the first 29 were used to compute the benchmarks.} The standard deviations were about 8 and 10 seconds, respectively. Now, we will now compare these measures with the STNN.

First, we consider \emph{batch processing}, referring to the computation of a batch of solutions with varying boundary conditions or problem parameters. This objective is relevant for the \emph{multi-query} use case mentioned in the introduction. On test group 1 (2647 problem inputs), the STNN required only 4.95 seconds on the GPU and 32.2 seconds on the CPU. By comparison, GMRES would need 3.1 hours on the GPU and 38 hours on the CPU. These values amount to a speedup of about 2250 and 4250, respectively.

Second, we consider single solution (sequential) processing, as might occur in a \emph{real-time} applications. Here, the speedup is less dramatic but still significant: on the GPU, the STNN took about 2.5 seconds, (1.8x faster than GMRES) while on the CPU it took 1.75 seconds (about 30x faster than GMRES).

\section{Discussion} \label{sec:discussion}

\subsection{Assessment and future directions} \label{sec:discussion-1}

Overall, the STNN is an effective reduced-order model for the parametric PDE problem in section \ref{sec:problem-statement}. The tensor networks likely play a major role, as they are highly effective reduced-order models of linear PDEs with \emph{fixed} parameters. Another factor is the preprocessing layer \eqref{eq:einsum-layer}, which regularizes the system to better represent the locality and symmetries of the PDE.

There are other, problem-specific features that may contribute to the generalizability of the STNN model. For example, as $\ell$ decreases, the boundary layer shrinks, and the solution in the bulk of the domain is increasingly dominated by the ``diffusion solution" $f_d$ from section \ref{sec:problem-statement}. It can be shown that the corresponding density $\rho_d(x,y)$ satisfies Laplace's equation in two dimensions \cite{Wagner2022}, which is a much simpler problem than the original PDE and one that is likely easier for the STNN to learn.\footnote{Note, however, that the complexity of the original PDE never disappears entirely. The main issue is the boundary conditions $g^{\pm}$, which are given for $f(x,y,w)$ instead of $\rho_d(x,y)$. While one can, in principle, derive effective boundary conditions for $\rho_d$, these depend nontrivially on $g^{\pm}$ and $\ell$.}

Because $\rho_d(x,y)$ is a 2D harmonic function, it is also interesting to ask whether there is a conformal mapping that relates problems with different values of the minor axis \cite{Morse1953}. Here, the issue of transforming boundary conditions prevents an analytic calculation. However, the question does raise the possibility that the generalization capability of the STNN derives, in part, from an approximate conformal mapping. In turn, this suggests that the STNN would generalize well to arbitrary curvilinear boundaries, provided that conformal mapping-based (body-fitted) grids are used \cite{Papamichael2010-td}.

Another way the STNN architecture could be enhanced is through the preprocessing layer (Eq. \eqref{eq:einsum-layer}). In the present implementation, this layer is essentially a 1D convolution in which the boundary nodes share the same weights. By contrast, a tiled convolutional layer \cite{Ngiam2010} would allow the weights to vary along the boundary, which may lead to a better description of complex curvilinear domains.

Finally, a challenge in applying the STNN to high-dimensional PDEs (e.g., $\geq 6$ independent variables) is the need to generate labeled training data by directly solving the PDE. Here, general strategies for \emph{reducing} the required quantity may be expedient \cite{Berg2018,Goswami2023}. For example, pretraining on coarse resolution simulations or simplified problem settings can accelerate convergence and reduce overfitting, as can the addition of regularization terms for problem symmetries and physical constraints. Another possibility is to pretrain individual model components (for example, through domain decomposition \cite{Seo2022}), and perform only lightweight training of the full, integrated model. In principle, both pretraining and regularization reduce the need for labeled data by improving the robustness of the training.

A more aggressive approach is to use the PDE residual directly as the loss, which could entirely eliminate the need for labeled data. This is the strategy adopted by physics-informed neural networks \cite{Raissi2019,Cuomo2022-ya,Goswami2023}. If some labeled data can be generated, then a hybrid approach may be most effective, in which the PDE residual serves as an additional regularization term \cite{Goswami2023}.

\subsection{Representational capacity} \label{sec:discussion-2}
In section \ref{sec:implementation-training}, I explained how the space of input functions (the boundary data) needs to be carefully defined. In particular, while we would like the STNN to handle arbitrary inputs, numerical evidence suggests this is not possible. For example, in section \ref{sec:results}, I showed two categories of inputs on which the STNN performed poorly: those with high spatial frequency and those producing boundary layer solutions. This notion has support on theoretical grounds: the tensor networks, like analogous matrix decompositions, derive their power partly by ignoring ``unimportant" modes. For example, a truncated singular value decomposition discards modes with large singular values, which are often the least important modes in physical applications. For PDEs with a low-dimensional solution manifold $\mathcal{M}$, it is reasonable to associate these discarded modes with solutions outside of $\mathcal{M}$.

If this picture is correct, the poor performance of the STNN on certain inputs is an unavoidable outcome of its reduced-order nature, which, absent a complete overhaul of the architecture, could not be ameliorated with hyperparameter tuning or more comprehensive training data. Even so, there are unanswered questions. In particular, for the problem considered here, the space of ``problematic inputs" is not unambiguously identified. While it appears to be related to the range of spatial correlations in the boundary data, it is not clear whether this is measured with respect to the grid resolution or an intrinsic length scale.

\vspace{3mm}

\textbf{Acknowledgments.} I thank the numerous developers of NumPy, SciPy, CuPy, TensorFlow, and t3f, without which this work would not have been possible.

\vspace{3mm}

\textbf{Data and code availability.} The relevant data and code are available on reasonable request to the author, and will be made public in a later version of this article.

\bibliography{bib}
\end{document}